\pdfoutput=1
\documentclass[10pt,twocolumn,letterpaper]{article}

\usepackage{iccv}
\usepackage{times}
\usepackage{epsfig}
\usepackage{graphicx}
\usepackage{amsmath}
\usepackage{amssymb}
\usepackage{multirow}
\usepackage{adjustbox}
\usepackage{subfigure}

\usepackage[pagebackref=true,breaklinks=true,letterpaper=true,colorlinks,bookmarks=false]{hyperref}

\iccvfinalcopy 

\def\iccvPaperID{****} 

\ificcvfinal\fi

\begin{document}

\title{ SPOTR: Spatio-temporal Pose Transformers for Human Motion Prediction }

\author{Avinash Ajit Nargund\\
University of California, Santa Barbara\\
{\tt\small anargund@ucsb.edu}
\and
Misha Sra\\
University of California, Santa Barbara\\
{\tt\small sra@cs.ucsb.edu}
}

\maketitle
\ificcvfinal\thispagestyle{empty}\fi

\begin{abstract}
3D human motion prediction is a research area of high significance and a challenge in computer vision. It is useful for the design of many applications including robotics and autonomous driving. Traditionally, autogregressive models have been used to predict human motion. However, these models have high computation needs and error accumulation that make it difficult to use them for realtime applications. In this paper, we present a non-autogressive model for human motion prediction. We focus on learning spatio-temporal representations non-autoregressively for generation of plausible future motions. We propose a novel architecture that leverages the recently
proposed Transformers. Human motion involves complex spatio-temporal dynamics with joints affecting the position and rotation of each other even though they are not connected directly. The proposed model extracts these dynamics using both convolutions and the self-attention mechanism. Using specialized spatial and temporal self-attention to augment the features extracted through convolution allows our model to generate spatio-temporally coherent predictions in parallel independent of the activity. Our contributions are threefold: (i) we frame human motion prediction as a sequence-to-sequence problem and propose a non-autoregressive Transformer to forecast a sequence of poses in parallel; (ii) our method is activity agnostic; (iii) we show that despite its simplicity, our approach is able to make accurate predictions, achieving better or comparable results compared to the state-of-the-art on two public datasets, with far fewer parameters and much faster inference.
\end{abstract}

 \section{Introduction}
Human motion prediction is the task of forecasting human poses conditioned on a sequence of observed poses. It is valuable in many applications including autonomous driving, animation, robotics, mixed reality, and healthcare. While humans can predict motions of other humans relatively easily to help them perform different tasks such as navigate through a crowd or play sports, the same is not true for algorithms. 

\begin{figure} [!t]
\centering
\includegraphics[width=\linewidth]{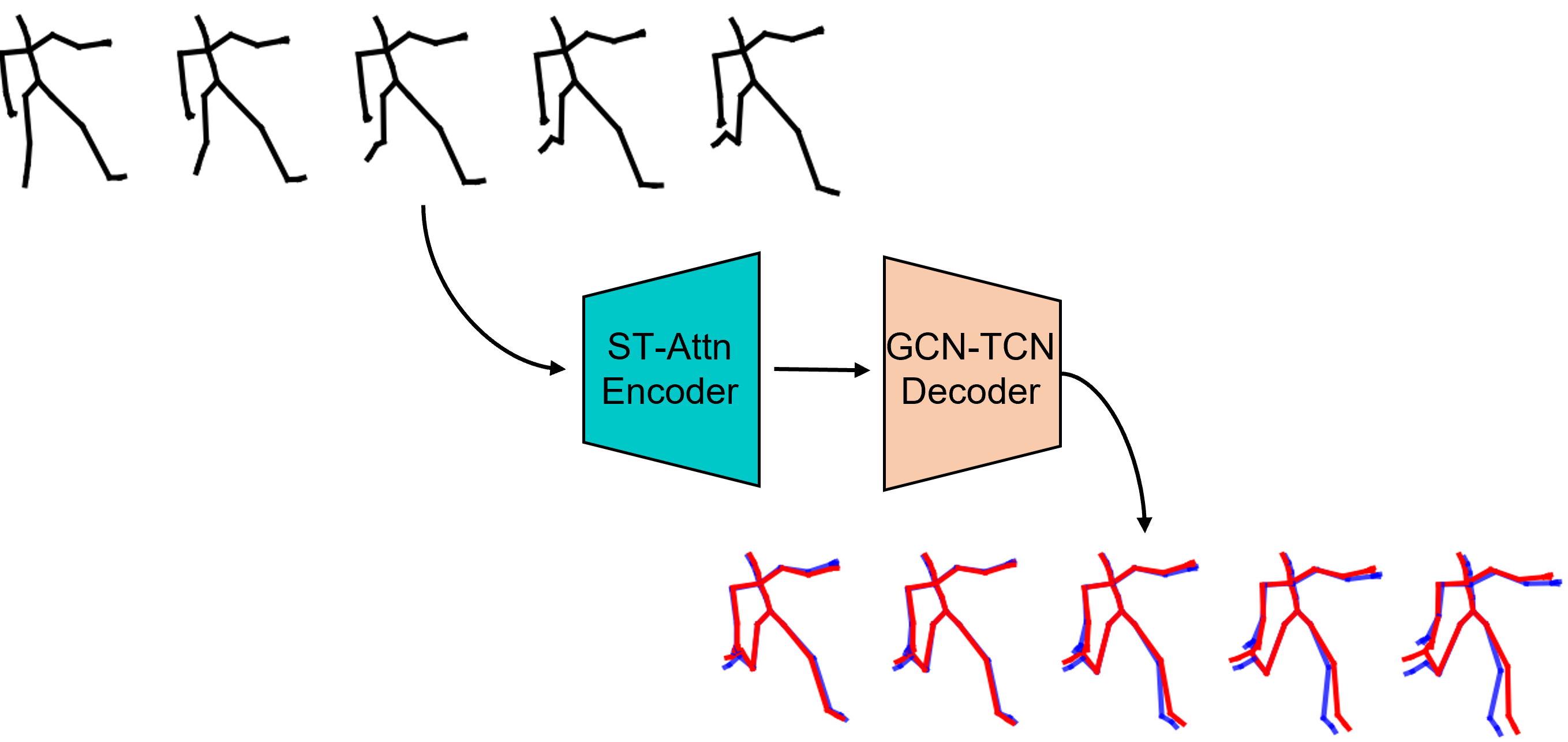}
\caption{Proposed approach for non-autoregressive motion prediction with a Transformer. Spatio-temporal feature extraction is followed by a temporal convolution in the encoder. Features generated by the encoder are passed through the decoder. To ensure the predicted quarternions represent valid rotations, we explicitly normalize them to be of unit length. All poses are predicted in parallel.}
\label{fig:illustration}
\end{figure}

Over the last decade, several attempts have been made to advance human motion forecasting. While earlier solutions have used Recurrent Neural Networks (RNNs) \cite{martinez2017human, wang2019imitation, pavllo2018quaternet} and Convolutional Neural Networks (CNNs) \cite{li2018convolutional}, models based on Graph Convolutional Networks (GCNs) \cite{dang_msr-gcn, li_multi_2021, hermes_application_2021} and Transformers \cite{aksan_spatio-temporal_2021, martinez-gonzalez_pose_2021} have become increasingly popular. Conventional methods relying on recurrent neural networks (RNNs) used stacks of LSTM or GRU modules and solve the task with autoregressive decoding, generating predictions sequentially, conditioned on previous predictions \cite{pavllo2018quaternet, ghosh2017learning, aksan_spatio-temporal_2021}. Autoregressive methods have two main shortcomings. First, the models are prone to accumulation of errors in prediction over time. This is because predictions are conditioned on previous predictions that already contain some error. Attempting to minimize these cumulative errors can eventually cause the predictions to collapse to a non-plausible static pose \cite{li2018convolutional,li2020multitask}. Second, autoregressive models are not parallelizable which makes it difficult to deploy these computationally intensive models in realtime interactive user-centered scenarios. Other motion prediction methods have included generative adversarial networks (GANs) \cite{hernandez2019human}, long short-term memory (LSTMs) \cite{ghosh2017learning}, and Markovian dynamics \cite{lehrmann2014efficient}. Most prior approaches have largely been replaced by deep learning methods fueled by the availability of large scale human motion datasets. However, 3D motion prediction remains a challenging task even though body joints and corresponding motions are highly correlated, the spatial relations and temporal evolution are difficult to model. 

In this work, we present a non-autogressive model architecture which explicitly considers the spatio-temporal aspects of human motion data for the 3D motion modeling task. Our approach is motivated by the recent success of Transformer models \cite{vaswani2017attention} on tasks such as
machine translation \cite{vaswani2017attention,camgoz2020sign}, music \cite{huang2018music}, animation \cite{huang2018music}, image captioning \cite{cornia2020meshed}, and image animation \cite{tao2022motion}. The original Transformer was designed for a one-dimensional sequences of words using self-attention \cite{vaswani2017attention,al2019character}. To use the self-attention mechanism for our inherently spatio-temporal 3D task of predicting human motion, we propose to decouple the temporal and spatial dimensions. Our model has an encoder-decoder architecture with the encoder extracting spatio-temporal features using self-attention augmented convolutions. The predictions are produced in a single pass by the decoder which is composed of multiple interleaved graph convolution and temporal convolution layers. 

In the proposed approach, we extract two sets of spatio-temporal features of input motion. One set of features is learned using a block of graph convolutions followed by convolution along the temporal dimension. The second set is obtained by adding the features obtained from the spatial and temporal attention blocks. In each input frame, temporal attention extracts the correlations between the positions of the same joint in the past frames while the spatial attention identifies the dependencies between the joints in the same frame. We concatenate the convolutional and self-attention features allowing the decoder to access and determine which segments of information are relevant for generating structurally and temporally coherent predictions. 

We evaluate our proposed model on the H3.6M \cite{ionescu2013human3} and CMU Mocap datasets in the short-term prediction setting. Our model matches the performance of state-of-the-art non-autoregressive model \cite{li_multi_2021} for short-term predictions using input over a much shorter time horizon (400ms vs 2000ms) compared to prior work \cite{li2020multitask, martinez-gonzalez_pose_2021}). Our approach allows for increased inference speed for time-sensitive applications, overcoming some of the limitations of autoregressive models. It is able to produce more accurate predictions with a relatively small number of parameters. Furthermore, in contrast with prior work \cite{martinez2021pose}, our method is activity-agnostic. 

\begin{figure*}[t]
  \centering
  \includegraphics[width=\linewidth]{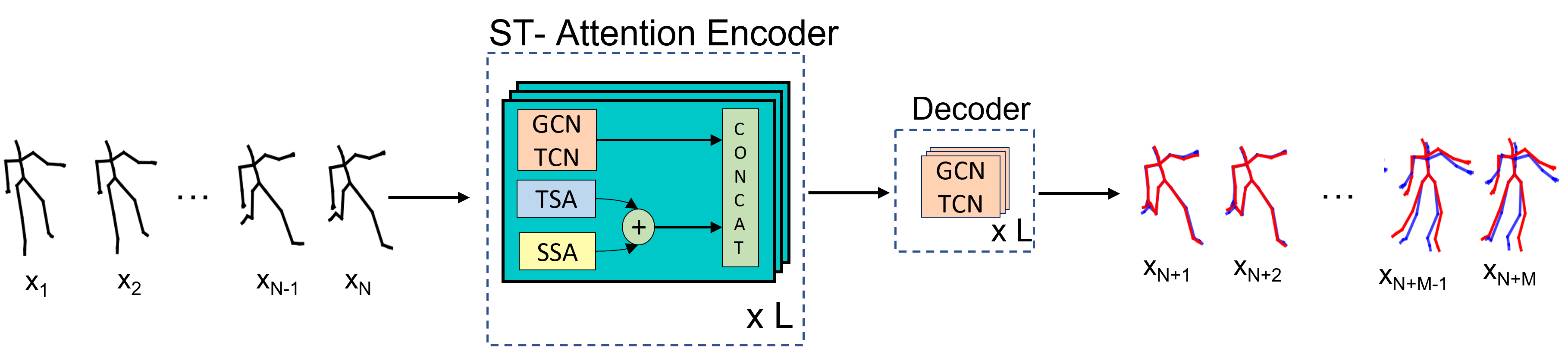}
  \caption{\textbf{Proposed Non-autoregressive Architecture.} Given an input motion sequence $X_{1:N}$ in quaternions we project it into an embedding space and add position encoding. The spatio-temporal features of the embeddings are extracted by the, (i) GCN-TCN block, and (ii) decoupled spatial (SSA) and temporal attention (TSA) blocks. We concatenate  the convolutional features with the sum of the attention features and pass them to the next layer. This concatenation allows the model to choose the most relevant features of the input motion. Finally, we generate the predictions in parallel using a 3-layer GCN-TCN decoder.}
  \label{fig:architecture}
\end{figure*}

\section{Related Work}
\label{sec:related_work}
Human pose forecasting is usually formulated as a sequence-to-sequence problem. A sequence of seed poses are used to extract features of the motion using an encoder and then the future poses are predicted, usually autoregressively, using a decoder. While earlier solutions were based on Recurrent Neural Networks (RNNs) \cite{martinez2017human, pavllo2018quaternet} or Convolutional Neural Networks (CNNs) \cite{li2018convolutional,hernandez2019human}, models based on Graph Convolutional Networks (GCNs) \cite{kipf2016semi} have become increasingly popular.

\subsection{Autoregressive Models} 
Aksan et. al. \cite{aksan_spatio-temporal_2021} proposed a Transformer-based architecture for the generative modelling of 3D human motion. They learn spatio-temporal dynamics autoregressively by using decoupled spatial and temporal self-attention. This enables their model to learn both structural and temporal dependencies explicitly and make accurate short-term predictions and also, generate possible future poses over long horizons. While the decoupled attention mechanisms are effective in learning the motion features, based on \cite{bello2019attention} we hypothesize that augmenting the attention features with spatio-temporal feature maps will improve the model performance. 

Multiscale GCNs are a popular choice of architecture for modelling human motion. GCNs operating on clusters of joints are used to learn the dynamics of the various joints. In \cite{dang_msr-gcn} the GCNs are used to extract features from fine to coarse scale and then from coarse to fine. The extracted multiscale features are then combined and decoded using a GCN decoder to obtain residuals between the input and the target pose sequence. \cite{li_multi_2021} fuse the extracted multiscale features across scales and feed them into a novel Graph Gated Recurrent Unit (G-GRU) to generate the predictions autoregressively.

Hermes et. al. \cite{hermes_application_2021} propose a low-complexity forecasting model based on the Graph-WaveNet. Each Graph-WaveNet block performs spatio-temporal convolutions followed by purely spatial graph convolutions. However, unlike the original Graph-WaveNet, they replace the full undirected skeletal graph with three directed graphs based on the joint kinematic tree. They use quaternions to represent the joints which have to be of unit length to be valid rotations. However the authors do not mention of how the validity of the predicted quaternions is ensured.  

Autoregressive models generate predictions one time step at a time which makes them slow and unsuitable for real-time applications. They also suffer from error accumulation and drift in the predictions \cite{martinez-gonzalez_pose_2021}. To avoid these pitfalls we propose a lightweight non-autogressive model which generates the predictions in parallel.

\subsection{Non-Autoregressive Models}
Development of non-autoregressive models can overcome some of the limitations of autogressive models as the goal to produce structurally and temporally coherent poses for realtime applications is crucial to many domains from mixed reality to pedestrian movements in autonomous driving scenarios.

The Discrete Cosine Transform (DCT) is popular choice for encoding the temporal features of the motion \cite{mao_learning_2020, mao_history_2020, cai_learning_2020}. Mao et. al. \cite{mao_history_2020}  use GCNs coupled with attention to predict the DCT coefficients which are then converted back into the time-domain. The idea of modeling the trajectories in the frequency domain is combined with the transformer architecture in \cite{cai_learning_2020}. However, they differ from traditional transformer-decoders by generating the predictions corresponding to a set of seed joints and then progressively estimating the locations of the other joints based on the kinematic chains of body skeletons. Models using the DCT to encode the temporal features are trained to predict both the input and target DCT coefficients to ensure continuity of the DCT coefficients. This additional computational burden might make it unsuitable for real-time applications. 

Martinez-Gonzalez et. al \cite{martinez-gonzalez_pose_2021} propose Pose Transformers to predict future poses. The joints are represented using Euler angles and the input and prediction sequences are projected from and to 3D pose vectors using GCNs. The projected input sequence is processed by a transformer encoder. The encoder output and a query sequence are used by the transformer decoder to compute the cross-attention which is used to generate  the predictions in one pass. A multi-task non-autoregressive motion prediction model is proposed in \cite{li2020multitask}. They represent the joints using quaternions and encode the seed skeleton sequence using a series of temporal and graph convolutional networks. The encoded contextual features are used to predict both the action and forecast the future motion. The future motion is predicted by combining them with positional embeddings and passing them through a decoder composed of GCN-TCN blocks. Both these methods however rely on the high-level action to guide the low-level predictions.

Different from prior work, our proposed approach operates on the joints in the temporal domain and learns the spatio-temporal dynamics from the sequence of input poses. Further, we generate the predictions using a shallow decoder independent of the action class of the motion.

\section{Background}
To make our manuscript self contained, we briefly introduce Spatial Temporal Graph Convolutional Networks (ST-GCNs) \cite{yan_spatial_2018} and Transformer self-attention \cite{vaswani2017attention}.

\subsection{Spatial Temporal Graph Convolutional Networks}
ST-GCNs proposed for skeleton-based action recognition are composed of a spatial convolution module followed by a temporal convolution module. Input to the ST-GCN is the joint quaternions on the graph nodes. Multiple layers of graph and temporal convolution are applied to gradually generate higher-level feature maps. The spatial features are computed for each partition of the adjacency matrix using the graph convolutional network (GCN) proposed in \cite{kipf2016semi}. Equation \ref{gcn} describes the operation of the GCN
\begin{equation}\label{gcn}
F_{out} = \sum_{i=1}^{K} F_{in} A_i W_i 
\end{equation}
where $K$ is the number of disjoint groups of $E_S$ defined in \ref{input} as the spatial edges, $A_i$ is the adjacency matrix of the $i^{th}$ partition. $F_{in}$ and $F_{out}$ are the input and the output feature maps.

\subsection{Transformer Self-Attention}
The original Transformer self-attention was proposed in \cite{vaswani2017attention} for Natural Language Processing (NLP) tasks. The self-attention mechanism is a sequence-to-sequence operation meant to augment the embedding of each word using the embeddings of the surrounding context. For each embedding  $e_i \in E = \{e_1, e_2, \dots, e_n \}$, a query $q \in \mathbf{R}^{d_q}$, a key $k \in \mathbf{R}^{d_k}$ and value $v \in \mathbf{R}^{d_v}$ is computed. The output embedding is computed as a weighted average of the values with the weights being determined with a dot-product between the query and keys. 

\begin{equation}
    Attention(Q, K, V) = softmax\left( \frac{Q K^T}{\sqrt{d_k}} \right) V
\end{equation}

where $Q, K, V$ are matrices containing the query, key and value vectors. In practice, a mechanism called the multi-head attention is used where the scaled dot-product attention is computed many times in parallel with different parameterized matrices and then combined to obtain the final embedding. This gives self-attention greater power of discrimination where different inputs can influence the output in different ways not possible in a single self-attention operation.

\section{Method}\label{sec:method}
This section describes our proposed approach. For an overview please refer to Figure~\ref{fig:architecture}. Our model consists of an attention augmented encoder and a simple decoder. The encoder effectively summarizes the spatio-temporal dynamics of the input motion by combining the convolutional and attention feature maps. The decoder uses only convolutions to generate the predictions in parallel.  


\subsection{Problem Formulation}\label{input}

Given a sequence of $N$ consecutive skeleton poses of a single human $X = \{x_1, x_2, \dots, x_N\}$ , we predict the next $M$ poses $X^{'} = \{x_{N+1}, x_{N+2}, \dots , x_{N+M}\}$. Each pose $x_i \in \mathbf{R}^{J \times 4}$ where $J$ is the number of joints and each joint is parameterized using quaternions. We represent the entire sequence of historical poses using a spatio-temporal graph \cite{yan_spatial_2018, plizzari_skeleton-based_2021}, , $G = (V, E)$. Here, $ V = \{ \nu_{ni} | n = 1, 2, \dots, N; i = 1, 2, \dots, J\}$ is the set of all nodes. Each node corresponds to a joint in a particular frame meaning the $\bigm| V \bigm| = N \times J$. $E$ is the set of all connections between the nodes. $E$ consists of two subsets - 
\begin{enumerate}
    \item Spatial Edges, $E_S = \{(\nu_{ni}, \nu_{nj}) | i, j = 1, 2, \dots, J ; n = 1, 2, \dots, N\}$ is the set of connections between pairs of joints $(i, j)$ at time $t$. This subset is further divided into $K$ disjoint groups using the partition strategies proposed in \cite{yan_spatial_2018}.
    \item Temporal Edges, $E_T = \{ (\nu_{ni}, \nu_{(n+1)i}) | i = 1, 2, \dots, J; n = 1, 2, \dots, N\}$ is the set of connections between consecutive time steps of a single joint $i$.
\end{enumerate} 

\begin{figure}[t]
\centering
\includegraphics[width=0.4\linewidth,scale=0.2]{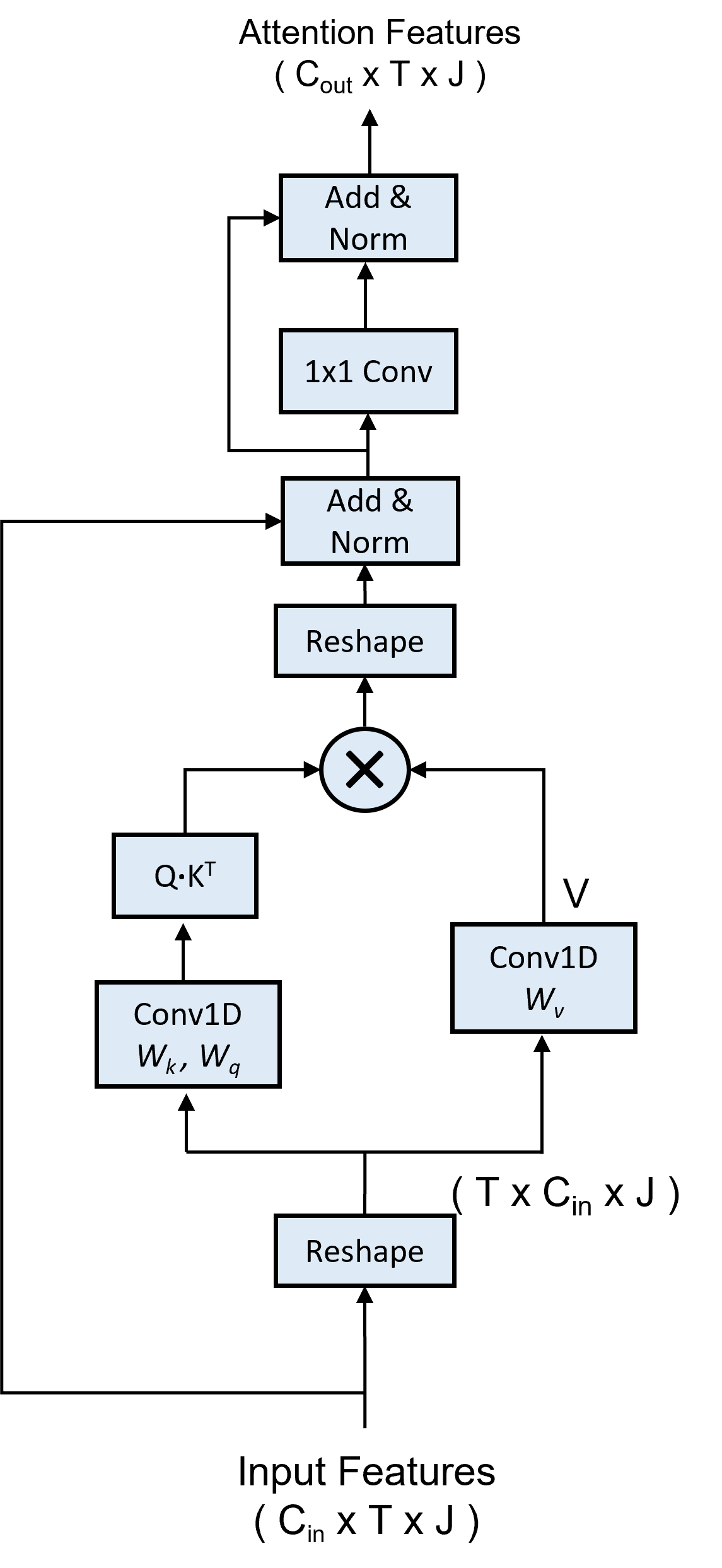}
\caption{Spatial Self-Attention Module}
\label{fig:ssa}
\end{figure}

\subsection{Spatio-temporal Pose Transformer}


Our proposed solution is shown in Figure \ref{fig:architecture}. The components of our model are - 
(i) \textbf{GCN-TCN-Unit}: The spatial and temporal features of the input motion history are extracted using graph convolution followed by temporal convolution 
(ii) \textbf{Spatial and Temporal Attention Modules}: Similar to other human motion analysis works \cite{aksan_spatio-temporal_2021, plizzari_skeleton-based_2021}, we extract the attention feature maps of the motion using spatial and temporal attention modules. 
%
%
(iii)\textbf{Decoder}: The decoder is made of a sequence of GCN-TCN units and it generates the predictions non-autoregressively. 

The input motion sequence is first projected into an embedding space using a $1x1$ convolution layer.

\subsection{GCN-TCN-Unit}
This is the spatio-temporal feature extraction block of our proposed model. The spatial features are extracted using the GCN proposed in \cite{kipf_neural_2018}. However instead of using a single adjacency matrix to represent the connections between the joints of the body we partition the adjacency using the partitioning strategies introduced in \cite{yan_spatial_2018}. The spatial feature map is then processed by a temporal convolution network (TCN). Dilated convolutions with kernels of size $1 \times T_s$ are used to extract the temporal features from the spatial feature maps.

\subsection{Spatial and Temporal Attention}

The self-attention mechanism is the prevalent choice for modeling long range dependencies in vision and sequence modeling tasks. Our model uses the traditional attention blocks but computes the self-attention along the spatial and temporal dimensions separately. 

\paragraph{Spatial Attention}
The spatial self-attention, as shown in Figure \ref{fig:ssa}, extracts the correlations between each pair of joints in each frame of the input sequence. Given the GCN-TCN  unit features corresponding to a particular joint $\nu_{ni}$ in frame $n$ , $f^{i}_{n} \in \mathbb{R}^{D_{in}}$. The spatial summary of the joints as a function of the other joints is computed using multi-head attention with $H$ heads. For each $\nu_{ni}$, a query vector $q^{n}_{i} \in \mathbb{R}^{d_q}$, a key vector  $k^{n}_{i} \in \mathbb{R}^{d_k}$ and value vector $v^{n}_{i} \in \mathbb{R}^{d_v}$ are computed by using linear transformations parameterized by trainable weights $\mathbf{W}_q \in \mathbb{R}^{D_{in} \times d_q } $, $\mathbf{W}_k \in \mathbb{R}^{D_{in} \times d_k } $ and $\mathbf{W}_v \in \mathbb{R}^{D_{in} \times d_v } $ that are shared across all the joints. Scaled dot-product attention $\alpha^{n}_{ij} = q^{n}_i {k^{n}_j}^T$ is used to compute the new feature vector $a_{i}^{n} \in \mathbb{R}^{D_{out}}$, 

\begin{equation} \label{ssa}
     a_{i}^{n} = \sum_{j} softmax_{j} \left( \frac{\alpha^{n}_{ij} }{\sqrt{d_k}} \right) v^{n}_{j}
\end{equation}
$H$ such new feature vectors are computed  and concatenated to obtain the output feature embedding for each $\nu_{ni}$, 
$\hat{f}^{i}_n =  concat \left( a_{1}^{n}, \dots, a_{H}^{n}\right)$.


\paragraph{Temporal Attention}
The temporal self-attention extracts the dependencies of each joint across all the frames. This assumes that features of associated with each joint is independent and the correlations between the poses in each frame with respect to one single joint. For the same joint $\nu_i$ from different frames $m$ and $n$, the query vector $q^{i}_m \in \mathbb{R}^{d_q}$ associated with $\nu_{mi}$, the key and value vectors - $k^{i}_n \in \mathbb{R}^{d_k}$ and $v^{i}_n \in \mathbb{R}^{d_v}$ associated with $\nu_{ni}$ are computed using linear trainable transformations similar to spatial attention.
With the correlation $\alpha^{i}_{mn} = q^{i}_m {k^{i}_n}^T$  the new feature vector is computed as 
\begin{equation} \label{tsa}
     a_{i}^{m} = \sum_{n} softmax_{n} \left( \frac{\alpha^{i}_{mn} }{\sqrt{d_k}} \right) v^{n}_{i}
\end{equation}
The resultant joint feature vector $a_{i}^{m} \in \mathbb{R}^{D_{out}}$ concatenated with the vectors from other attention heads to obtain the output feature embedding.  

\paragraph{}
 We pass the spatial and temporal attention features through a $1\times1$ convolution layer to ensure that the attention map can be concatenated with the convolutional features along the feature dimension.  The output of the spatial and temporal attention blocks are normalized and summed to obtain the final attention features. It is then augmented by concatenating the spatio-temporal convolutional features and passed to the next layer.   




  

\begin{figure*}
\begin{minipage}{.25\textwidth}
  \centering
  \includegraphics[scale=0.3]{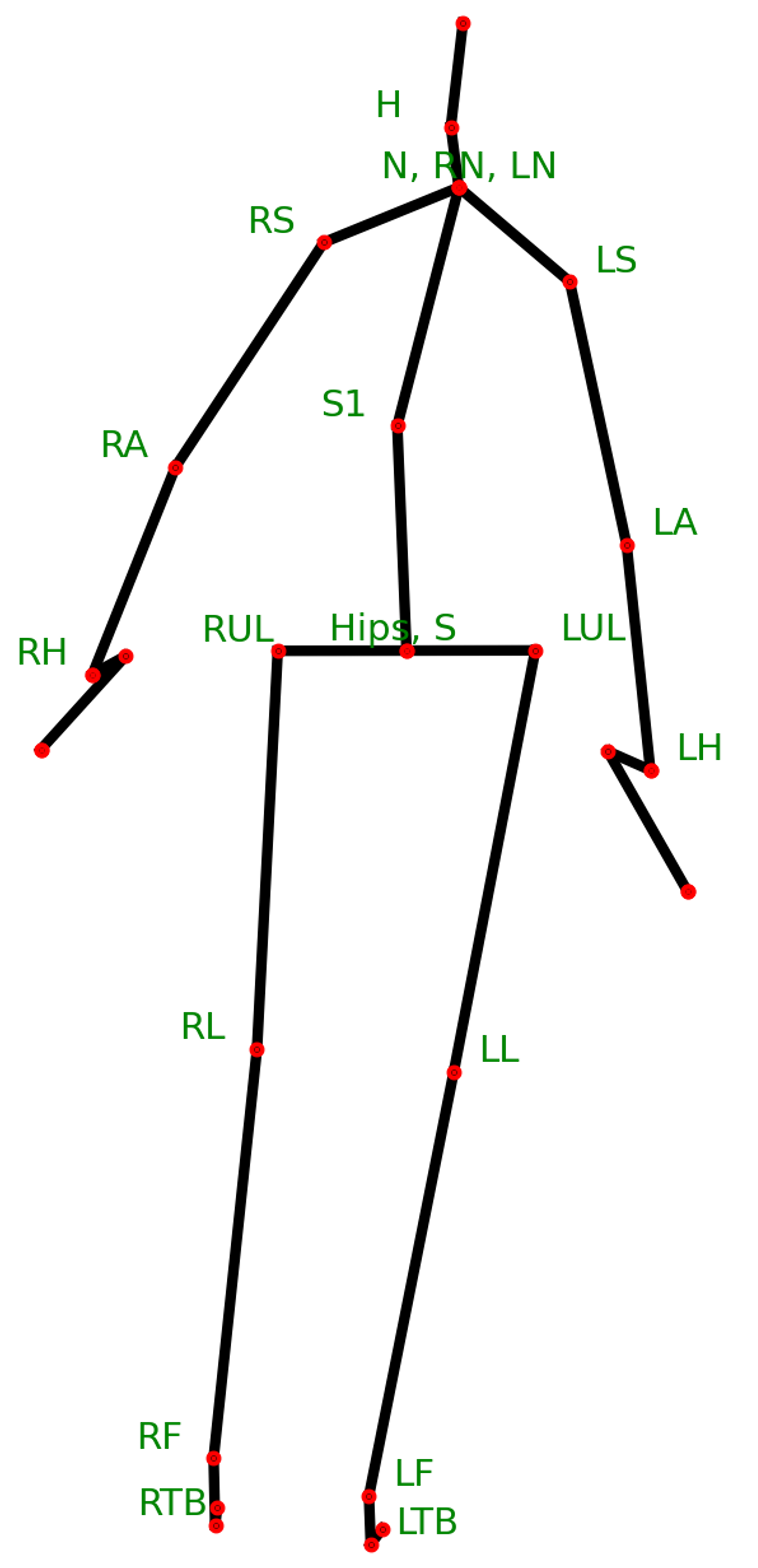}
\end{minipage}%
\begin{minipage}{.8\textwidth}
  \subfigure{\includegraphics[ width=\linewidth]{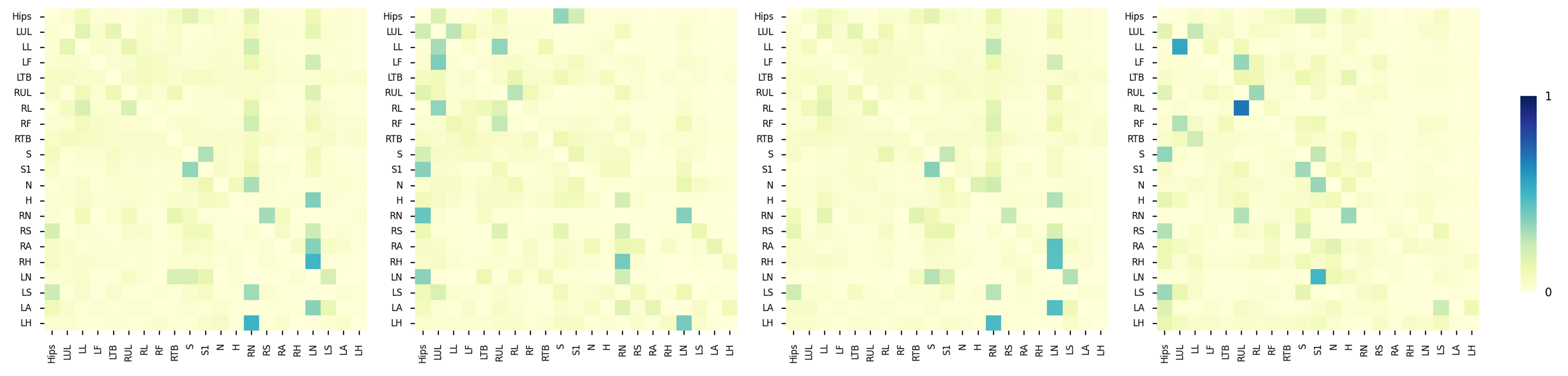} }\\
    \subfigure{\includegraphics[width=\linewidth]{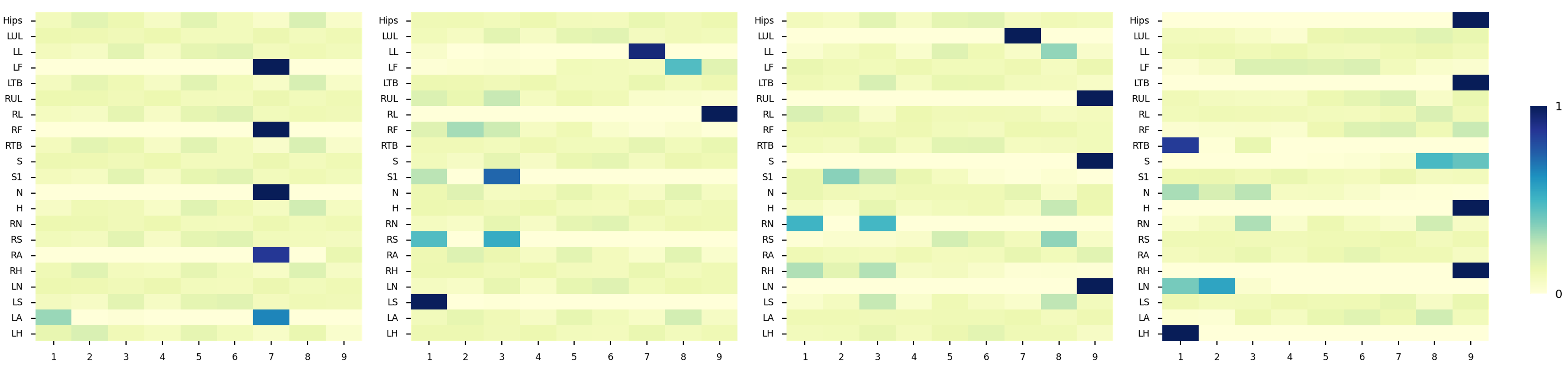} }%

\end{minipage}
 \caption{The mean spatial attention weights (top row) and the temporal attention weights (bottom row) associated with the last input frame given 10 frames of \textit{walkingtogether} from Human3.6M.  From the 4 attention heads visualized we can see that the model is able to attend to joints that are not connected in the kinematic chain and is able to extract the dependencies between the spatio-temporal features of each joint from the past frames.}
     \label{fig:attn_weights}
\end{figure*}



\subsection{Decoder}
The predictions are generated by the passing the features generated by the encoder through three GCN-TCN blocks. To ensure the predicted quaternions represent valid rotations we explicitly normalize them to be of unit length.

\subsection{Training}

We train the model by using a loss function composed of weighted mean distance in Euler angle space \cite{pavllo2018quaternet} and mean Euclidean distance between joints in 3D coordinates. Consider the predicted quaternions, $\hat{X}$ associated with a single training sample $X$. It is the converted to joint Euler angles, $\hat{X}^e$ and 3D positions of the joints, $\hat{X}^p$ using forward kinematics. The loss is computed as,   

\begin{equation} \label{loss_function}
    \mathcal{L}\left( \mathbf{X}, \mathbf{\hat{X}} \right) = \alpha * E + \beta * P 
\end{equation}
where, 
\begin{equation*}
    E = \frac{1}{N \times J} \sum_{n=0}^{N} \sum_{j=0}^{J}   \bigm| \left\{ \left( x_{n,j}^{e} - \hat{x^e}_{n,j} + \pi \right) \bmod 2\pi \right\} - \pi \bigm|,
\end{equation*}
\begin{equation*}
    P = \frac{1}{N \times J} \sum_{n=0}^{N} \sum_{j=0}^{J} \lVert x_{n,j}^{p} - \hat{x^p}_{n,j} \rVert_{2}
\end{equation*}

and
$\alpha, \beta$ are scalars used to weigh the contribution of the individual losses.

\begin{table*}
\centering
\begin{adjustbox}{width=1\textwidth}
\begin{tabular}{l *{4}{c}| *{4}{c} | *{4}{c} | *{4}{c} }
    \multirow{2}{*}{\textbf{interval (ms)}} &
      \multicolumn{4}{c}{\textbf{Basketball}} &
      \multicolumn{4}{c}{\textbf{Basketball Signal}} &
      \multicolumn{4}{c}{\textbf{Directing Traffic}}&
      \multicolumn{4}{c}{\textbf{Jumping}}
      \\
    & 80 & 160 & 320 & 400 & 80 & 160 & 320 & 400 & 80 & 160 & 320 & 400& 80 & 160 & 320 & 400\\
    \hline
    LTD \cite{mao_learning_2020} & \textbf{0.33} & 0.52 & 0.89 & 1.06 & \textbf{0.11} & \textbf{0.2} & 0.41 & 0.53 & \textbf{0.15} & \textbf{0.32} & \textbf{0.52}  & \textbf{0.6}  & \textbf{0.31} & \textbf{0.49} & \textbf{1.23}  & 1.39 \\

    mNAT \cite{li2020multitask} & 0.34 & 0.49 & \textbf{0.86} & \textbf{1.01} & 0.15 & 0.24 & 0.48 & 0.61 & 0.20 & 0.41 & 0.65  & 0.77  & 0.38 & 0.56 & 1.29  & 1.45\\
    \hline
    Ours  & 0.37 & 0.53 & \underline{0.88} & \underline{1.05} & 0.17 & \underline{0.22} & \textbf{0.37} & \textbf{0.47} & 0.36 & 0.53 & 0.7  & 0.84  & 0.76 & 0.9 & \underline{1.28}  & \textbf{1.3}\\
    \hline
    \hline

\multirow{2}{*}{\textbf{interval (ms)}} &
      \multicolumn{4}{c}{\textbf{Running}} &
      \multicolumn{4}{c}{\textbf{Soccer}} &
      \multicolumn{4}{c}{\textbf{Walking}} &
      \multicolumn{4}{c}{\textbf{Wash Window}} \\
    & 80 & 160 & 320 & 400 & 80 & 160 & 320 & 400 & 80 & 160 & 320 & 400 & 80 & 160 & 320 & 400\\
\hline 
    LTD \cite{mao_learning_2020} & 0.33 & 0.55 & 0.73 & 0.74 & \textbf{0.18} & \textbf{0.29} & 0.61 & \textbf{0.71} & 0.33 & 0.45 & 0.49  & 0.53  & \textbf{0.22} & 0.33 & 0.57  & 0.75\\

    mNAT \cite{li2020multitask} & \textbf{0.24} & 0.43 & 0.53 & \textbf{0.56} & 0.20 & 0.33 & \textbf{0.59} & 0.72 & 0.31 & 0.37 & 0.40  & 0.46  & 0.23 & 0.36 & 0.66  & 0.86\\
    \hline
    Ours & \underline{0.31} & \textbf{0.38} & \textbf{0.5} & \underline{0.57} & 0.35 & 0.45 & 0.68 & 0.80 & \textbf{0.24} & \textbf{0.28} & \textbf{0.34}  & \textbf{0.36}  & \textbf{0.22} & \textbf{0.31} & \textbf{0.53}  & \textbf{0.63}\\
    \hline
    \hline
  \end{tabular}
\end{adjustbox}

\begin{adjustbox}{width=0.3\textwidth}
\begin{tabular}{l *{4}{c} }
    \multirow{2}{*}{\textbf{interval (ms)}} &
      \multicolumn{4}{c}{\textbf{Average}}
      \\
    & 80 & 160 & 320 & 400  \\
    \hline
    LTD \cite{mao_learning_2020} & \textbf{0.25} & \textbf{0.39} & 0.68 & 0.79  \\
    mNAT \cite{li2020multitask} & 0.26 & 0.4 & 0.68 & 0.8 \\
    \hline
    Ours  & 0.34 & 0.45 & \textbf{0.67} & \textbf{0.74} \\
    \hline
    \hline
\end{tabular}%
\end{adjustbox}%

\caption{Comparing the performance of our model on euler angle error (lower is better) with the existing state-of-the-art models on the CMU Mocap dataset. Our model outperforms both of them on longer time horizons indicating that it is able to learn the complex spatio-temporal dynamics of joints while playing sports as well. }\label{cmu_ee}%
\end{table*}%

\begin{table*}
\centering
\begin{adjustbox}{width=1\textwidth}
\begin{tabular}{l *{4}{c}| *{4}{c} | *{4}{c} | *{4}{c} }
    \multirow{2}{*}{\textbf{interval (ms)}} &
      \multicolumn{4}{c}{\textbf{Basketball}} &
      \multicolumn{4}{c}{\textbf{Basketball Signal}} &
      \multicolumn{4}{c}{\textbf{Directing Traffic}}&
      \multicolumn{4}{c}{\textbf{Jumping}}
      \\
    & 80 & 160 & 320 & 400 & 80 & 160 & 320 & 400 & 80 & 160 & 320 & 400& 80 & 160 & 320 & 400 \\
    \hline
    LTD \cite{mao_learning_2020} & \textbf{14.0} & \textbf{25.4} & 49.6 & 61.4 & \textbf{3.5} & \textbf{6.1} & \textbf{11.7} & \textbf{15.2} & \textbf{7.4} & \textbf{15.1} & 31.7  & 42.2  & \textbf{16.9} & \textbf{34.4} & 76.3  & 96.8 \\

    \hline
    Ours  & 17.5 & 26.0 & \textbf{45.1} & \textbf{57.0} & 8.3 & 9.5 & 15.10 & 19.5 & 11.5 & 15.7 & \textbf{26.8}  & \textbf{34.8}  & 35.0 & 45.7 & \textbf{70.1}  & \textbf{86.0}\\
    \hline   

\multirow{2}{*}{\textbf{interval (ms)}} &
      \multicolumn{4}{c}{\textbf{Running}} &
      \multicolumn{4}{c}{\textbf{Soccer}} &
      \multicolumn{4}{c}{\textbf{Walking}} &
      \multicolumn{4}{c}{\textbf{Wash Window}} \\
    & 80 & 160 & 320 & 400 & 80 & 160 & 320 & 400 & 80 & 160 & 320 & 400 & 80 & 160 & 320 & 400\\
\hline 
    LTD \cite{mao_learning_2020} & 25.5 & 36.7 & 39.3 & 39.9 & \textbf{11.3} & \textbf{21.5} & \textbf{44.2} & \textbf{55.8} & \textbf{7.7} & 11.8 & 19.4  & 23.1  & \textbf{5.9} & \textbf{11.9} & \textbf{30.3}  & \textbf{40.0}\\

    \hline
    Ours & \textbf{15.7} & \textbf{19.3} & \textbf{25.9} & \textbf{31.3} & 21.5 & 30.4 & 48.0 & 56.1 & 8.3 & \textbf{10.9} & \textbf{15.5}  & \textbf{16.6}  & 13.6 & 19.6 & 33.6  & 40.8    \\
    \hline
    
  \end{tabular}
\end{adjustbox}

\begin{adjustbox}{width=0.3\textwidth}
\begin{tabular}{l *{4}{c} }
    \multirow{2}{*}{\textbf{interval (ms)}} &
      \multicolumn{4}{c}{\textbf{Average}}
      \\
    & 80 & 160 & 320 & 400  \\
    \hline
    LTD \cite{mao_learning_2020} & \textbf{11.5} & \textbf{20.4} & 37.8 & 46.8  \\

    \hline
    Ours  & 16.2 & 21.9 & \textbf{34.6} & \textbf{42.4} \\
    \hline
    \hline
\end{tabular}%
\end{adjustbox}%

\caption{Performance comparison with state-of-the-art non-autoregressive models in terms of the MPJPE (in millimeters) on the CMU Mocap dataset. Our model outperforms the existing models for longer time-horizons indicating that the attention-augmented spatio-temporal convolution is capable of capturing long term dynamics from the input motion. }\label{cmu_mpjpe}%
\end{table*}%

\begin{table*}
\centering
\begin{adjustbox}{width=1\textwidth}
\begin{tabular}{l *{4}{c}| *{4}{c} | *{4}{c} | *{4}{c} | *{4}{c} }
    \multirow{2}{*}{\textbf{interval (ms)}} &
      \multicolumn{4}{c}{\textbf{Walking}} &
      \multicolumn{4}{c}{\textbf{Eating}} &
      \multicolumn{4}{c}{\textbf{Smoking}}&
      \multicolumn{4}{c}{\textbf{Discussion}} &
      \multicolumn{4}{c}{\textbf{Directions}} 
      \\
    & 80 & 160 & 320 & 400 & 80 & 160 & 320 & 400 & 80 & 160 & 320 & 400& 80 & 160 & 320 & 400 & 80 & 160 & 320 & 400\\
    \hline

    POTR \cite{martinez-gonzalez_pose_2021} & \textbf{0.16} & 0.40 & 0.62 & 0.73 & \textbf{0.11} & \textbf{0.29} & 0.53 & 0.68 & \textbf{0.14} & \textbf{0.39} & 0.84  & 0.82  & \textbf{0.17} & 0.56 & 0.85  & 0.96 & \textbf{0.20} & 0.45 & 0.79  & 0.91\\

    mNAT \cite{li2020multitask} & 0.17 & \textbf{0.29} & \textbf{0.45} & \textbf{0.53} & 0.17 & 0.31 & \textbf{0.48} & \textbf{0.54} & 0.22 & 0.4 & \textbf{0.81}  & \textbf{0.78}  & 0.23 & \textbf{0.54} & \textbf{0.72}  & \textbf{0.8} & 0.27 & \textbf{0.43} & \textbf{0.58}  & \textbf{0.67}\\
    \hline
    Ours  & 0.21 & \underline{0.32} & \underline{0.54} & \underline{0.6} & 0.21 & 0.35 & 0.58 & 0.69 & 0.28 & 0.45 & 0.85  & 0.84  & 0.34 & 0.69 & 0.97  & \underline{0.92} & 0.31 & 0.51 & \underline{0.77}  & \underline{0.84}\\
    \hline
    \hline

\multirow{2}{*}{\textbf{interval (ms)}} &
      \multicolumn{4}{c}{\textbf{Greeting}} &
      \multicolumn{4}{c}{\textbf{Phoning}} &
      \multicolumn{4}{c}{\textbf{Posing}} &
      \multicolumn{4}{c}{\textbf{Purchases}} &
      \multicolumn{4}{c}{\textbf{Sitting}} \\
    & 80 & 160 & 320 & 400 & 80 & 160 & 320 & 400 & 80 & 160 & 320 & 400 & 80 & 160 & 320 & 400 & 80 & 160 & 320 & 400\\
\hline 

    POTR \cite{martinez-gonzalez_pose_2021} & \textbf{0.29} & 0.69 & 1.17 & 1.30 & 0.5 & 1.1 & 1.5 & 1.65 & 0.18 & 0.52 & 1.18  & 1.47  & \textbf{0.33} & 0.63 & 1.04  & 1.09 & \textbf{0.25} & 0.47 & 0.92  & 1.09\\

    mNAT \cite{li2020multitask} & 0.33 & \textbf{0.51} & \textbf{0.79} & \textbf{0.94} & 0.53 & \textbf{0.92 }& \textbf{1.15} & \textbf{1.28} & \textbf{0.18} & \textbf{0.38} & \textbf{0.81}  & \textbf{1.00}  & 0.4 & \textbf{0.55} & \textbf{0.85}  & \textbf{0.89} & 0.29 & \textbf{0.46} & 0.84  & 1.04\\
    \hline
    Ours & 0.4 & \underline{0.64} & \underline{1.05} & \underline{1.25} & \textbf{0.46} & \underline{0.97} & \underline{1.39} & \underline{1.51} & 0.31 & 0.64 & \underline{1.17}  & \underline{1.34}  & 0.51 & 0.75 & 1.05  & 1.15 & 0.33 & 0.48 & \textbf{0.82}  & \textbf{1.03}\\
    \hline
    \hline

\multirow{2}{*}{\textbf{interval (ms)}} &
      \multicolumn{4}{c}{\textbf{Sitting Down}} &
      \multicolumn{4}{c}{\textbf{Taking Photo}} &
      \multicolumn{4}{c}{\textbf{Waiting}} &
      \multicolumn{4}{c}{\textbf{Walking Dog}} &
      \multicolumn{4}{c}{\textbf{Walking Together}} \\
    & 80 & 160 & 320 & 400 & 80 & 160 & 320 & 400 & 80 & 160 & 320 & 400 & 80 & 160 & 320 & 400 & 80 & 160 & 320 & 400\\
\hline 

    POTR \cite{martinez-gonzalez_pose_2021} & \textbf{0.25} & \textbf{0.63} & 1.00 & 1.12 & \textbf{0.12} & 0.41 & 0.71 & 0.86 & \textbf{0.17} & 0.56 & 1.14  & 1.37  & \textbf{0.35} & 0.79 & 1.21  & 1.33 & \textbf{0.15} & 0.44 & 0.63  & 0.70\\

    mNAT \cite{li2020multitask} & 0.34 & 0.66 & \textbf{0.94} & \textbf{1.06} & 0.14 & \textbf{0.32} & \textbf{0.54} & \textbf{0.68} & 0.22 & \textbf{0.48} & \textbf{0.79}  & \textbf{0.97}  & 0.43 & \textbf{0.7} & \textbf{0.86}  & \textbf{1.03} & 0.16 & \textbf{0.28} & \textbf{0.51}  & 0.62\\
    \hline
    Ours & 0.56 & 0.79 & \underline{0.99} & \underline{1.08} & 0.23 & 0.43 & \underline{0.62} & \underline{0.75} & 0.28 & \underline{0.51} & \underline{0.9}  & \underline{1.12}  & 0.45 & 0.83 & 1.22  & 1.39 & 0.21 & \underline{0.39} & \underline{0.54}  & \textbf{0.59} \\
    \hline
  \end{tabular}
\end{adjustbox}

\begin{adjustbox}{width=0.3\textwidth}
\begin{tabular}{l *{4}{c} }
    \multirow{2}{*}{\textbf{interval (ms)}} &
      \multicolumn{4}{c}{\textbf{Average}}
      \\
    & 80 & 160 & 320 & 400  \\
    \hline
     POTR \cite{martinez-gonzalez_pose_2021} & 0.23 & 0.55 & 0.94 & 1.08 \\

    mNAT \cite{li2020multitask} & \textbf{0.27} & \textbf{0.48} & \textbf{0.74} & \textbf{0.85}  \\
    \hline
    Ours  & 0.33 & 0.58 & \underline{0.89} & \underline{1.00} \\
    \hline
    \hline
\end{tabular}%
\end{adjustbox}%

\caption{Comparing the performance of our model on euler angle error (lower is better) with the existing non-autoregressive models in all 15 motion categories of the Human3.6M dataset in terms of Euler angles. Our lightweight model consistently performs on par (underlined values) with the much more computationally-intensive models in multiple human motion categories. }\label{h36res}
\end{table*}%

\begin{figure*}
    \centering
        \subfigure[\textit{walkingtogether}]{\includegraphics[scale=0.425]{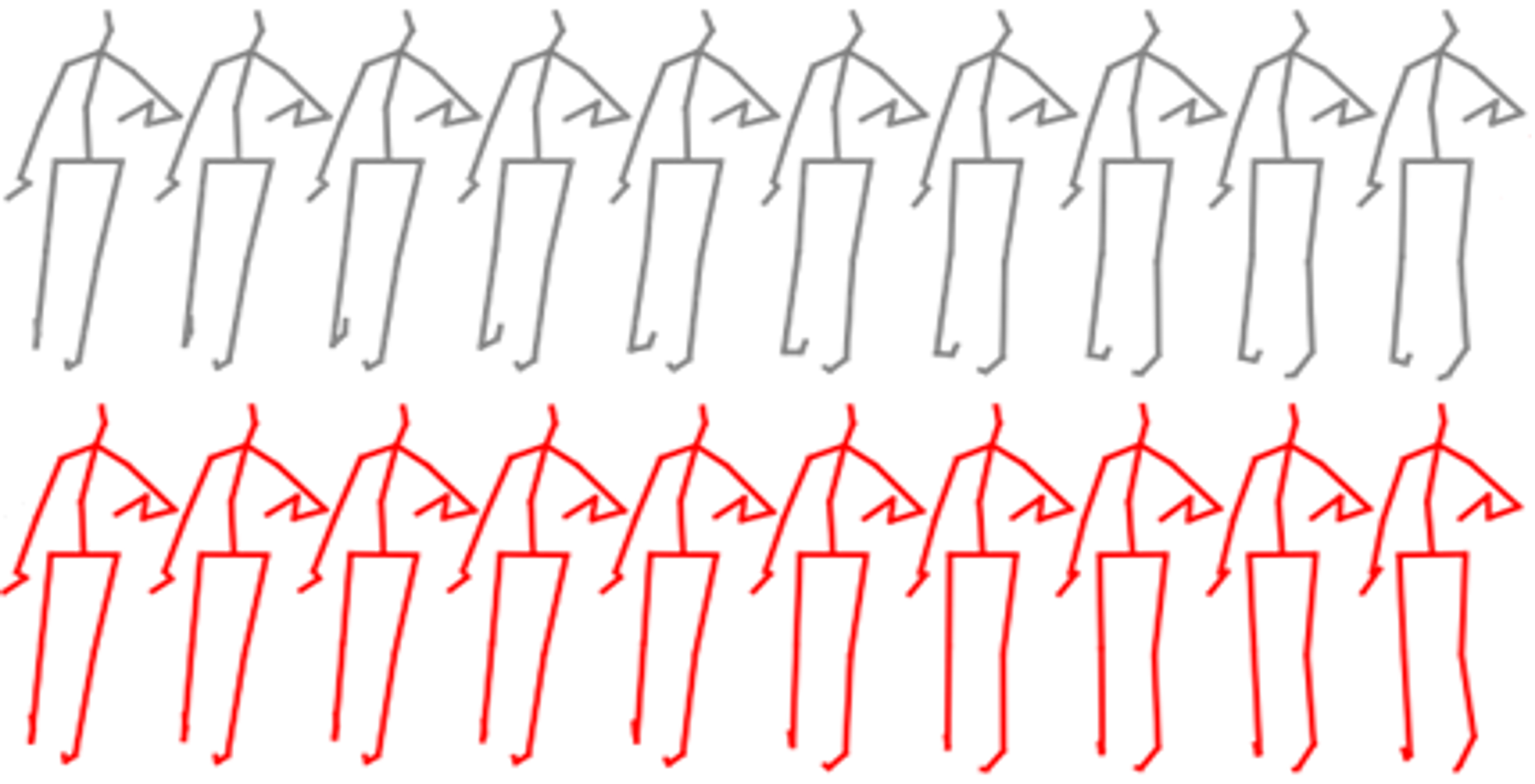}}
        \hspace{5mm}
        \subfigure[\textit{walking}]{\includegraphics[scale=0.425]{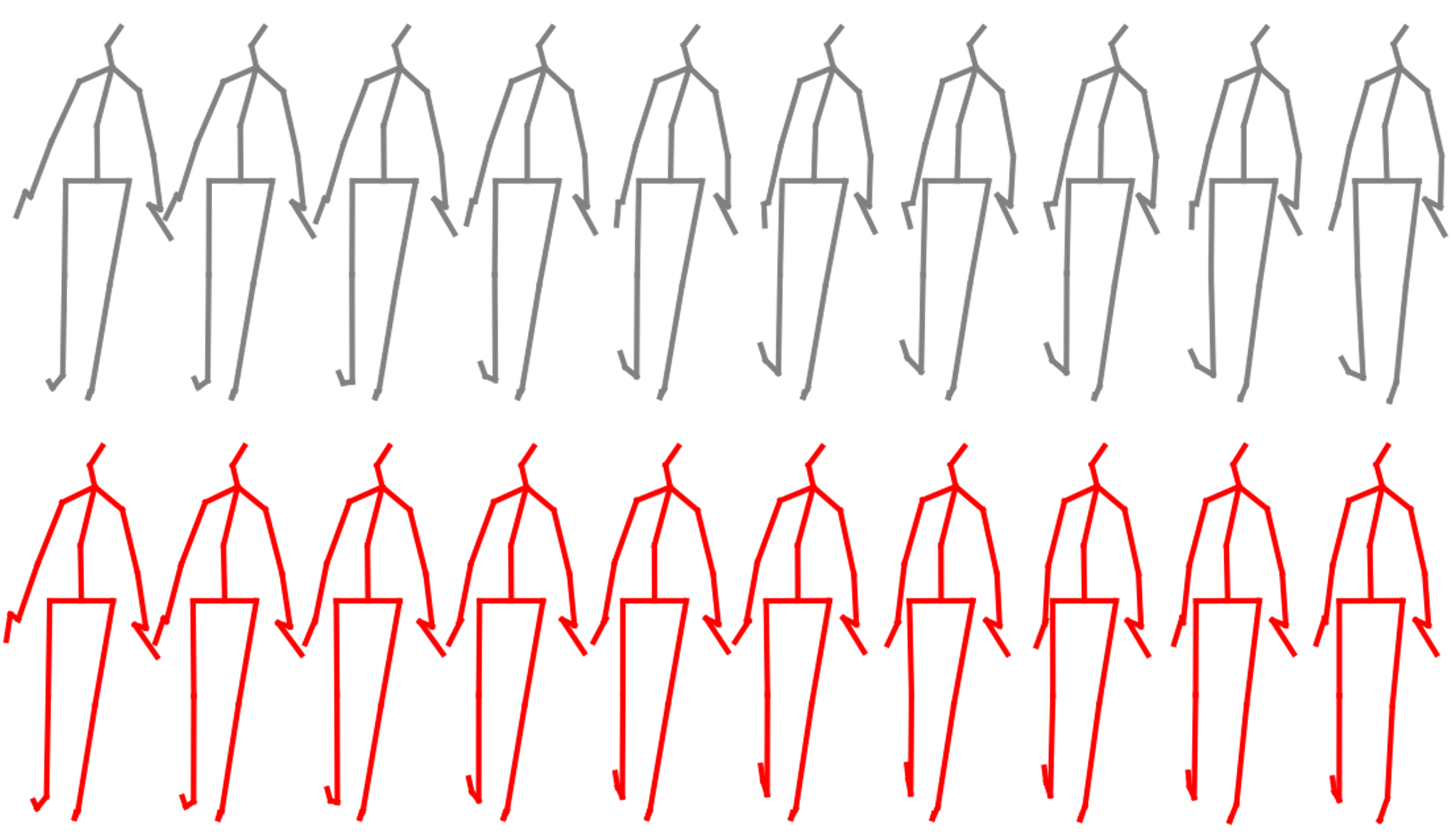}}
        \caption{Qualitative results on \textit{walkingtogether} and \textit{walking} samples from Human3.6 from 10 frames of input. We can see that the predictions (in red) match closely with the ground truth (in grey). Note particularly the match in the more challenging joints such as hands and feet. }
\end{figure*}

\begin{figure*}
    \centering
        \subfigure[soccer]{\includegraphics[scale=0.39]{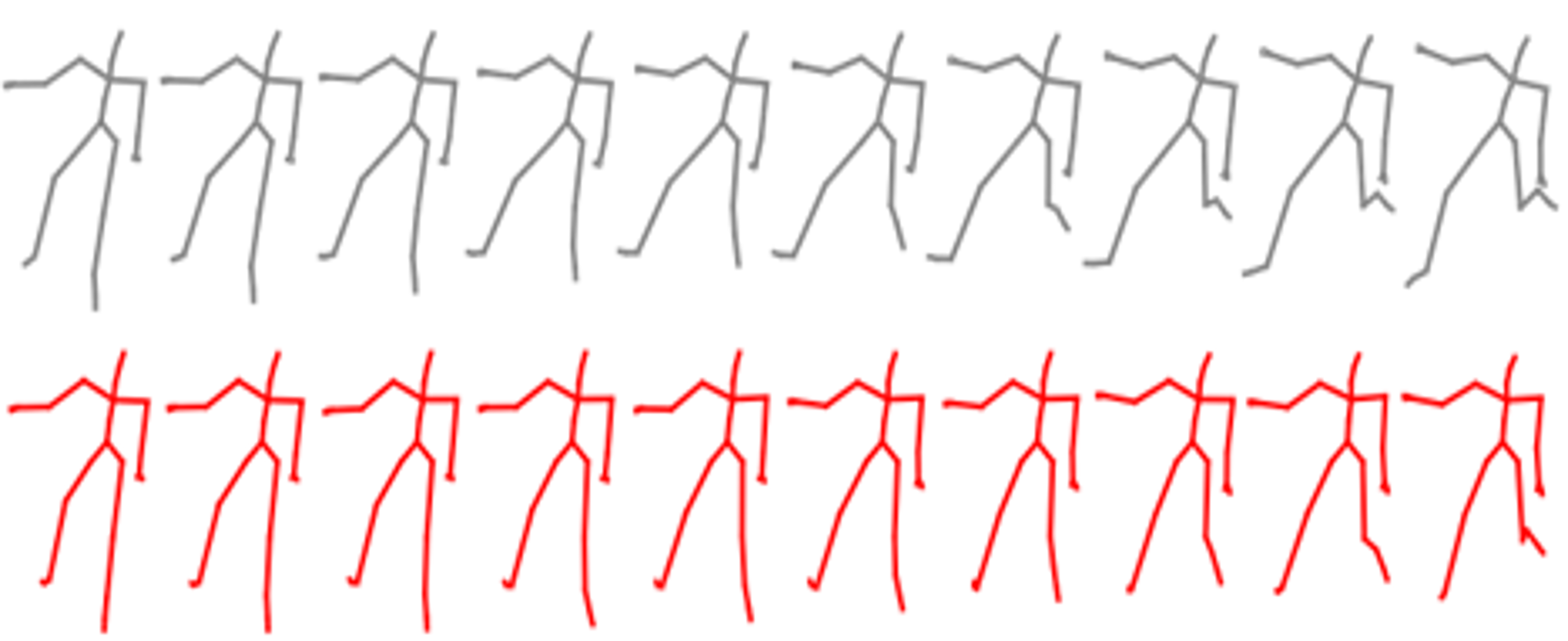}} 
        \subfigure[washwindow]{\includegraphics[scale=0.39]{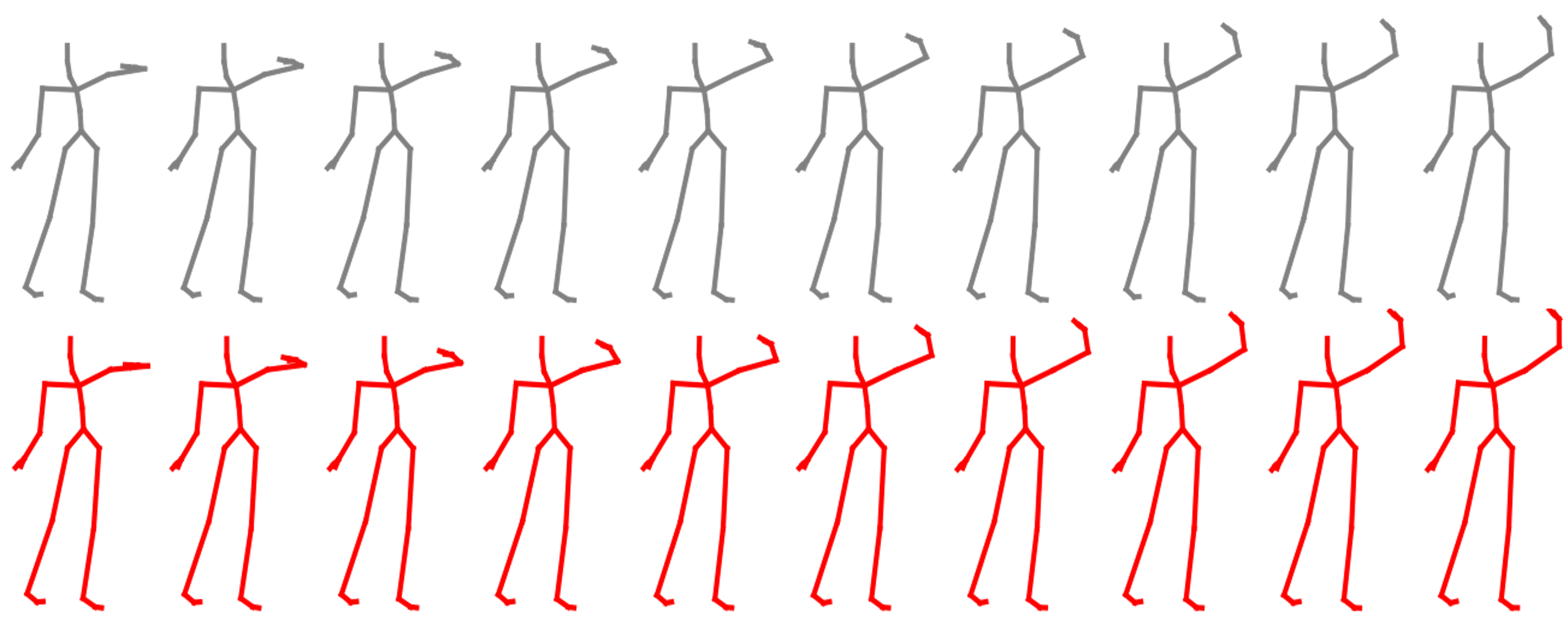}}
        \caption{Qualitative results on \textit{soccer} and \textit{washwindow} samples from CMU Mocap from 10 frames of input. Despite being activity-agnostic, in the \textit{soccer} sample we can see that the spatio-temporal dynamics in the predictions (in red) are similar to that of the ground truth (in grey). We also see the model does well in predicting parts of periodic motion in the \textit{washwindow} sample. }
\end{figure*}%

\section{Experiments}

We compare our model with the state-of-the-art methods quantitatively. We evaluate our model on the CMU Mocap \cite{cmu_mocap} and Human 3.6M \cite{ionescu2013human3} datasets. In the section we introduce the datasets, the evaluation metrics and report results in both error in joint angles and 3D positions. 

\subsection{Datasets}
\paragraph{Human3.6M}
The Human3.6M dataset is the most popular human motion dataset used for benchmarking the pose prediction task. It contains 3.6 million poses and each pose is represented using a skeleton composed of 32 joints. We remove joints with a constant rotation resulting in 21 joints in the skeleton .It consists of 15 actions such as walking, phoning, and eating performed by 7 subjects. The global translation and rotation in the pose sequences which are provided in axis-angle format are first removed \cite{mao_history_2020, li2020multitask}. Then they are downsampled to 25 frames per second and  converted into a sequence of quaternions. The data from subject 11 is used to tune hyperparameters and the model is tested on data from subject 5. Following \cite{pavllo2018quaternet}, we augment the dataset by mirroring all the pose sequences.

\paragraph{CMU Motion Capture}
The CMU-Mocap dataset contains motion data of humans performing various actions such running, walking, and jumping. The poses are represented using a skeleton composed of 38 joints. For fair comparison, we train on data associated with the same actions as in \cite{mao_learning_2020, li2020multitask}. The pose sequences are pre-processed similar to the sequences in Human3.6M dataset.  %

\subsection{Implementation Details}%
Our network is implemented in Pytorch. We use the spatial partitioning strategy from \cite{yan_spatial_2018} to partition the adjacency matrix representing the joint connections. Dilations of 1, 2, and 4 are used in the temporal convolution layers in the GCN-TCN unit in order increase their receptive field. The network is trained using the RAdam optimizer \cite{liu2019variance} as it eliminates the need for learning rate warmup during training. We use learning rate of $3 \times 10^{-4}$ and decay it by a factor of $0.99$ every epoch. The model is regularized using a weight decay of $1e-5$ and a dropout rate of $0.1$ in the convolutional layers operating along the temporal dimension. By performing some test runs we picked the values $\alpha =10$ and $\beta=0.1$. In the SSA unit shown in \ref{fig:architecture} we use a temporal convolution layer after the spatial attention. Analogous to this we use a GCN unit to extract the spatial features before extracting temporal attention features. We use relative position encoding \cite{bello2019attention} in both the attention blocks. %
\subsection{Metrics}%
We use the evaluation procedure used in previous works \cite{mao_history_2020, li_multi_2021} and report the short term prediction results ($80 - 400 ms$) in both the Euler angle space and 3D positions. We use the average Euclidean distance in the Euler angle space and the Mean Per Joint Position Error (MPJPE) in millimeters. The errors for a single frame, $n$ are computed as,%
\[
    E = \lVert x^e_{n} - \hat{x}^e_{n} \rVert_2 
\]%
\[
    MPJPE = \frac{1}{J} \sum_{j=1}^{J} \lVert x^p_{n,j} - \hat{x}^p_{n,j} \rVert_2
\]

\subsection{Results}
The performance of our model is provided in Table \ref{cmu_ee}, \ref{cmu_mpjpe}, and \ref{h36res}. For the Human3.6M dataset, to ensure a fair comparison we compare our model performance only to the existing non-autoregressive works \cite{li2020multitask, martinez-gonzalez_pose_2021} which operate on the input motion sequence in the time domain. From Table \ref{h36res} we observe that errors in our model in the joint angle space are almost on par with the state-of-the-art non-autoregressive models despite being activity-independent and despite having far fewer parameters. 

Our model uses 6 GCN-TCN blocks and 2 sets of temporal and spatial self-attention modules. This is extremely light in comparison to, 
(a) \cite{li2020multitask} which uses 6 GCN-TCN blocks in its encoder and decoder along with an additional activity recognition network, and (b) POTR \cite{martinez-gonzalez_pose_2021} which uses 4 layers in the encoder and decoder each made of GCNs and multi-head attention modules. In addition, the POTR model \cite{martinez-gonzalez_pose_2021} uses the last pose as the query in the encoder-decoder attention and hence performs well in the immediate time horizon (80-160ms) on average while our model performs better than POTR in predicting for longer time horizons (320-400ms). All these models use 2000ms of input motion history to generate the predictions. This is five times more than the number of input frames we use. These factors highlight the efficacy of using both convolutions and the self-attention mechanism to learn the spatio-temporal dynamics of human motion. Our results demonstrate that it is possible to build a lightweight model that is performative and  that can be used in realtime interactive human-centered applications.%

The work by \cite{li2020multitask} is the only non-autoregressive work evaluated on the CMU mocap dataset but they too do not provide metrics in terms of MPJPE. Hence, we include \cite{mao_learning_2020}, which uses DCTs to model the temporal variation, in our comparison. From Table \ref{cmu_ee} and \ref{cmu_mpjpe} we see that out model outperforms the state-of-art models on CMU Mocap for the longer time horizons (320 and 400ms). This indicates that our lightweight model is capable of learning the spatio-temporal dynamics associated with complex sports actions.

\section{Conclusion}

We introduce a novel lightweight spatio-temporal transformer (SPOTR) network for the 3D human motion prediction task. We show that augmenting spatio-temporal convolution with self-attention is very effective in making human motion predictions. Our non-autoregressive approach mitigates the issues of error accumulation, non-parallelizability, and static poses seen in autoregressive models while still providing comparable performance with a small number of parameters, significantly shorter seed sequences and faster computation time. We also demonstrate that the attention mechanism can be used to attain insights about the model's behavior. Finally, our lightweight model allows us to integrate build realtime interactive human-centered applications to solve real world problems.


{\small
\bibliographystyle{ieee_fullname}
\bibliography{main}
}

\end{document}